\documentclass{article}
\usepackage{ijcai22}

\usepackage{times}
\usepackage{url}
\usepackage[hidelinks]{hyperref}
\usepackage[utf8]{inputenc}
\usepackage[ruled]{algorithm2e} 
\usepackage{graphicx}
\usepackage[round]{natbib}
\usepackage{subcaption}
\usepackage{amsmath}
\usepackage{amsfonts}
\usepackage{amssymb}
\usepackage{amsthm}
\usepackage{booktabs}
\usepackage{wrapfig}
\title{How to Train Your Program: \\
a Probabilistic Programming Pattern
for Bayesian Learning From Data}

\author{
	David Tolpin
	\affiliations
  Ben-Gurion University of the Negev, Israel
  PUB+
  \emails
  david.tolpin@gmail.com
}

\begin{document}

\maketitle
\begin{abstract}
	We present a Bayesian approach to  machine learning with
	probabilistic programs. In our approach, training on
	available data is implemented as inference on a hierarchical
	model. The posterior distribution of model parameters is
	then used to \textit{stochastically condition} a
	complementary model, such that inference on new data yields
	the same posterior distribution of latent parameters
	corresponding to the new data as inference on a hierarchical
	model on the combination of both previously available and
	new data, at a lower computation cost. We frame the approach
	as a design pattern of probabilistic programming referred to
	herein as `stump and fungus', and evaluate realization of
	the pattern on synthetic and real-world case studies.
\end{abstract}

\section{Introduction}

The ultimate Bayesian approach to learning from data is embodied
by hierarchical models~\citep{GCS+13,GT+16,M20}. In a
hierarchical generative model, the distribution of each
observation $y_{ij}$ from the $i$th group of observations
depends on group parameter $\theta_i$, and the distribution of
each $\theta_i$ depends on hyperparameter $\tau$:
\begin{equation} 
	\begin{aligned}
	\tau & \sim H \\
	\theta_i|\tau & \sim D(\tau) \\
	y_{ij}|\theta_i & \sim F(\theta_i)
	\end{aligned}
	\label{eqn:hier}
\end{equation}
A hierarchical model can be thought of as a way of inferring, or
`learning', the prior of $\theta_i$ from all observations
in the data set. Consider the following example problem:
multiple boxes are randomly filled by $K$ marbles from a bag
containing a mixture of blue and white marbles. We are presented
with a few draws with replacement from each of the boxes, $y_{ij}$
being the $j$th draw from the $i$th box; our goal is to infer
the number of blue marbles $\theta_i$ in each box.  Intuitively,
since the boxes are filled from the same bag, the posterior
distribution of $\theta_i$ should account both for draws from
the $i$th box and, indirectly, for draws from all other boxes.
This is formalized by the following hierarchical model:
\begin{equation}
	\begin{aligned}
		\tau & \sim \mathrm{Beta}(1, 1) \\
		\theta_i|\tau & \sim \mathrm{Beta}(K\tau, K(1 - \tau)) \\
		y_{ij}|\theta_i & \sim \mathrm{Bernoulli}(\theta_i)
	\end{aligned}
	\label{eqn:marbles}
\end{equation}

Model~\eqref{eqn:marbles} learns from the data in the sense that
inference for each box is influenced by draws from all boxes.
However, learning from \textit{training data} to improve
inference on \textit{future data} with a  hierarchical model is
computationally inefficient --- if a new box is presented, one
has to add observations of the new box to the previously
available data and re-run inference on the extended data set.
Inference performance can be improved by employing data
subsampling~\citep{KCW14,BDH14,BDH17,MA14,QVK+18}, but the whole
training data set still needs to be kept and made accessible to the
inference algorithm.  A hierarchical model cannot `compress', or
summarize, training data for efficient inference on future
observations.

An alternative approach, known as \textit{empirical
Bayes}~\citep{R51,C85,R92}, consists in adjusting the hyperprior
based on the training data; e.g.  by fixing $\tau$ at a likely
value,  or by replacing $H$
in~\eqref{eqn:hier} with a suitable approximation of the
posterior distribution of $\tau$. However, empirical Bayes is
not Bayesian. While practical efficiency of empirical Bayes was
demonstrated in a number of settings~\citep{R92}, in other
settings empirical Bayes may result in a critically misspecified
model and overconfident or biased inference outcomes.

In this work, we propose an approach to learning from data in
probabilistic programs which is both Bayesian in nature and
computationally efficient. First, we state the problem of
learning from data in the context of Bayesian generative models
(Section~\ref{sec:problem}).  Then, we introduce the
approach and discuss its implementation
(Section~\ref{sec:idea}).  For evaluation, we apply the
approach to inference in synthetic and real-world problems
(Section~\ref{sec:studies}). Finally, we
conclude with a overview of related work and discussion
(Sections~\ref{sec:related} and~\ref{sec:discussion}).

\section{Problem: Learning from Data}
\label{sec:problem}

The challenge we tackle here is re-using inference outcomes on
the training data set for inference on new data. Formally,
population $\mathcal{Y}$ is a set of sets $\pmb{y}_i \in \mathcal{Y}$ of
observations $y_{ij} \in \pmb{y}_i$.  Members of each $\pmb{y}_i$ are
assumed to be drawn from a known distribution $F$ with
unobserved parameter $\theta_i$, $y_{ij} \sim F(\theta_i)$.
$\theta_i$ are assumed to be drawn from a common distribution
$H$. Our goal is to devise a scheme that, given a subset $Y
\subset \mathcal{Y}$, the \textit{training set}, infers the
posterior distribution of $\theta_k|Y, \pmb{y}_k$ for any
$\pmb{y}_k \in \mathcal{Y}$ in a shorter amortized time than
running inference on a hierarchical model $Y \cup \{\pmb{y}_k\}$.  By
amortized time we mean here average time per $\pmb{y}_k,\,k \in 1:K$
as $K \to \infty$.

In other words, we look for a scheme that works in two stages.
At the first stage, inference is performed on the training set
$Y$ only. At the second stage, the inference outcome of the
first stage is used, together with $\pmb{y}_k$, to infer $\theta_k|Y,
\pmb{y}_k$. We anticipate a scheme that `compresses' the training set
at the first stage, resulting in a shorter running time of the
second stage. Such scheme bears similarity to the conventional
machine learning paradigm: an expensive computation on the
training data results in shorter running times on new data.

\section{Main Idea: Stump and Fungus}
\label{sec:idea}

In quest of devising such a scheme, we make two
observations which eventually help us arrive at a satisfactory
solution:
\begin{enumerate}
	\item In Bayesian modelling, information about data 
		is usually conveyed through conditioning of the
		model on various aspects of the data.
	\item In a hierarchical model, influence of the $i$th group of
		observations on the hyperparameters $\tau$ and, consequently,
		on other groups, passes exclusively through the group parameters
		$\theta_i$.
\end{enumerate}

\begin{wrapfigure}{r}{0.33\linewidth}
	\includegraphics[width=\linewidth]{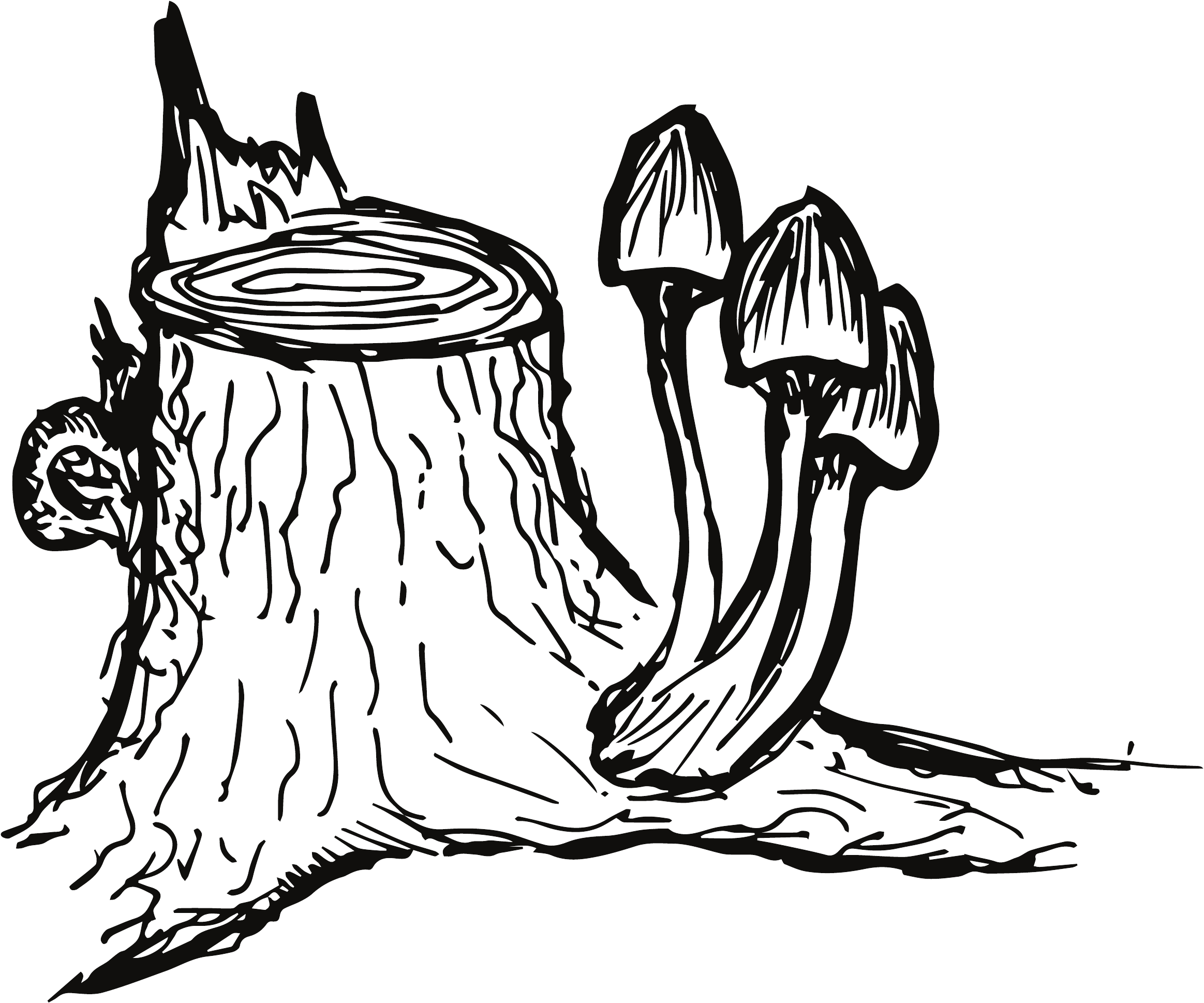}
\end{wrapfigure}

If, instead of conditioning on training data $y_i$, we could
condition on parameters $\theta_i$ corresponding to the training
data, then we could perform inference on new data item $y_k$ at
a lower time and space complexity.  Continuing the well known
analogy  between a hierarchical model and a tree, with the
hyperparameter $\tau$ at the root and observations $y_{ij}$ in
the leaves, we can liken a model which receives all $\theta_i$
of the training data and new data item $y_k$ to a \textit{stump}
--- the hierarchical model with the trunk cut off just after the
hyperparameters --- and a \textit{fungus} growing on the stump
--- the new data item.  The problem is, of course, that we infer
\textbf{distributions}, rather than fixed values, of $\theta_i$,
and the model must be, somewhat unconventionally, conditioned on
the distributions of $\theta_i$.

However, a recently introduced notion of stochastic
conditioning~\citep{TZR+21} makes conditioning on distributions
of $\theta_i$ possible, both theoretically and in the practical
case when the posteriors of $\theta_i$ are approximated using
Monte Carlo samples. Stochastic conditioning~\citep{TZR+21} extends
deterministic conditioning $p(x\vert y=y_0)$,
i.e.~conditioning on some random variable $y$
taking on a particular value $y_0$, to conditioning $p(x\vert y
\sim D_0)$ on $y$ having the marginal distribution $D_0$.
A probabilistic model with stochastic conditioning is a tuple
$(p(x, y), D)$ where $p(x, y)$ is the joint probability density
of random variable $x$ and observation $y$ and $D$  is the
distribution of observation $y$, with density $q(y)$. To infer $p(x\vert y \sim
D)$, one must be able to compute $p(x, y \sim D)=p(x)p(y \sim D\vert x)$. $p(y
\sim D \vert x)$ has the following unnormalized density:
\begin{equation}
    p(y \sim D\vert x) = \exp \left( \int_Y (\log p(y\vert x))\,q(y)dy \right)
    \label{eqn:prob-D-given-x}
\end{equation}

Conditioning the model both \textit{stochastically} on the
posterior distributions of $\theta_i$ on training data and
\textit{deterministically} on new data $y_k$ would yield the
same posterior distribution of $\theta_k$ as inference on the
full hierarchical model. However, that would also mean that the
inference algorithm can sample from the posterior distribution of 
$\theta_i$ to infer $y_k$. To achieve the objective of
`compression' of the training set for faster inference on new
data, the parameter posterior can be approximated by a weighted
finite set of samples. This is related to Bayesian
coresets~\citep{HCB16,ZKK+21}, although with essential differences,
discussed in Section~\ref{sec:related}.  Based on this, we
propose the `stump-and-fungus' pattern for learning from data in
probabilistic programs:
\begin{enumerate}
	\item Training is accomplished through inference on a
		hierarchical model, in the usual way. 
	\item The parameter posterior is approximated by a set of
		$M$ weighted samples $(\tilde {\pmb{\theta}}, \pmb{w})$.
		\begin{itemize}
		\item $\tilde\theta_i$ are drawn from the mixture of
			parameter posteriors of all groups. The
			set size $M$ is chosen as a compromise between
			approximation accuracy and inference performance.
		\item The weights are selected to minimize approximation
			error of the hyperparameter posterior, as detailed in
			Section~\ref{sec:formal}.
		\end{itemize}
	\item For inference on new data $\pmb{y}$, a stump-and-fungus
		model is employed:
		\begin{equation}
			\begin{aligned}
				\tau & \sim  H \\
				(\tilde {\pmb{\theta}}, \pmb{w})|\tau & \sim D(\tau)\quad \mbox{ --- stochastic conditioning} \\
				\theta |\tau & \sim D(\tau) \\
				\pmb{y}|\theta &  \sim F(\theta)
			\end{aligned}
			\label{eqn:stump-and-fungus}
		\end{equation}
		The unnormalized conditional log probability of $(\tilde
		{\pmb{\theta}}, \pmb{w})$ given $\tau$ is computed
		as~\eqref{eqn:p-theta-w}, following~\cite{TZR+21}:
		\begin{equation}
			\log p((\tilde {\pmb{\theta}}, \pmb{w})|\tau) = \sum_{j=1}^M w_j \log p(\tilde \theta_j|\tau) 
			\label{eqn:p-theta-w}
		\end{equation}
\end{enumerate}
A gain in time and space complexity follows from replacing the
dataset $Y$ with a weighted sample set $(\tilde{\pmb{\theta}},
\pmb{w})$ of size $M$. Many inference algorithms scale at least
linearly with the size of the data, hence the corresponding
complexity term decreases from $O(|Y|)$ to $O(M)$. In addition, 
the likelihood of a weighted sample skips the lowest level of
the hierarchy and is thus cheaper to compute than
of an observation, though the exact gain depends on the
particular model.

\begin{figure*}
	\begin{subfigure}{0.245\linewidth}
		\centering
		\includegraphics[width=0.95\linewidth]{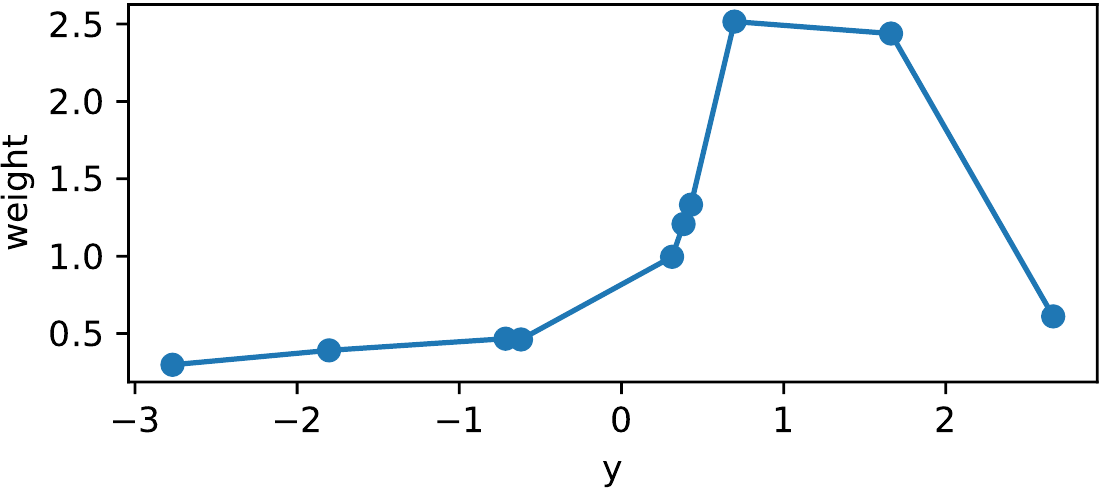}
		\caption{weighted sample set}
		\label{fig:normal-surrogate}
	\end{subfigure}
	\begin{subfigure}{0.745\linewidth}
		\centering
		\includegraphics[width=0.95\linewidth]{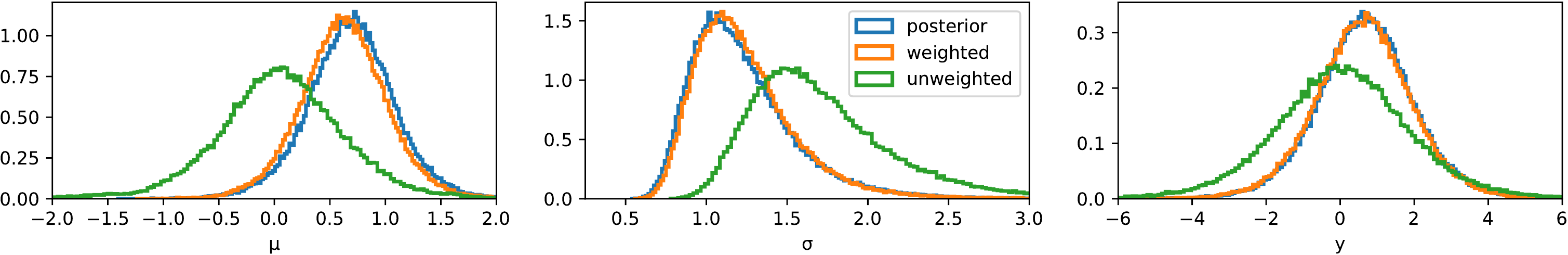}
		\caption{posterior}
		\label{fig:normal-posterior}
	\end{subfigure}
	\caption{Inferring parameters of a normal distribution}
	\label{fig:normal}
\end{figure*}

Although two models  --- hierarchical and stump-and-fungus ---
are involved in the pattern,  the models are in fact two roles
fulfilled by the same generative model, combining stochastic
conditioning on training data and deterministic conditioning on
new data (consisting potentially of multiple data items).  This
preserves a common practice in machine learning in which the same
model is used for both training and inference.

\section{Weighted Sample Set Approximation of Parameter Posterior}
\label{sec:formal}

In Section~\ref{sec:idea}, we suggested to stochastically
condition the model on the posterior distributions of $\theta_i$
of the training data. We further suggested that, to `compress'
the training set, the distribution be approximated by a finite
weighted sample set $(\tilde {\pmb{\theta}}, \pmb{w})$ of size
$M$. Let us assume that the sample set $(\tilde{\pmb{\theta}})$
is drawn and fixed, and show how the weights $\pmb{w}$ are
computed.

Our objective is to approximate the posterior of the hyperparameter
$\tau$ given the training set $Y$, by conditioning $\tau$ on
$(\tilde {\pmb{\theta}}, \pmb{w})$. We can achieve this by
selecting weights $\pmb{w}$ such that the KL divergence between
$p(\tau|Y)$ and $p(\tau|(\tilde {\pmb{\theta}}, \pmb{w}))$ is
minimized. Since $p(\tau|Y)$ does not depend on $\pmb{w}$,
minimizing $KL(p(\tau|Y)||p(\tau|(\tilde {\pmb{\theta}},
\pmb{w}))$ is equivalent to the following
maximization problem:
\begin{equation}
	\begin{aligned}
		\pmb{w} & = \arg \max_{\pmb{w}} S(\pmb{w}) \\
		S(\pmb{w}) & = \int_{\tau \in T} p(\tau|Y) \log p(\tau|(\tilde {\pmb{\theta}}, \pmb{w})) d\tau  \\
		\mbox{where, from \eqref{eqn:p-theta-w}} & \\
		p(\tau|(\tilde {\pmb{\theta}}, \pmb{w})) & =  \!\!\frac {p(\tau) \exp \left(\sum_{j=1}^M w_j \log p(\tilde\theta_j|\tau)\right)} {\!\!\int_{\tau' \in T} p(\tau') \exp \left(\sum_{j=1}^M w_j \log p(\tilde\theta_j|\tau')\right)d \tau'}
	\end{aligned}
	\label{eqn:w-arg-max-p-log-q}
\end{equation}
In the context of Bayesian inference in hierarchical generative models, the
posterior is commonly approximated by a set of Monte Carlo samples. Assuming
that $\tau|Y$ is approximated by a (large) set of $N$ samples $\tau_1 ... \tau_N$, 
$S(\pmb{w})$ can be estimated as
\begin{equation}
	\begin{aligned}
		\hat S(\pmb{w}) &= \sum_{i=1}^N \log \frac {p(\tau_i) \exp \left(\sum_{j=1}^M w_j \log p(\tilde\theta_j|\tau_i)\right)} {\sum_k p(\tau_k) \exp \left(\sum_{j=1}^M w_j \log p(\tilde\theta_j|\tau_k)\right)} \\
		& = \sum_{i=1}^N \left(\log p(\tau_i) + \sum_{j=1}^M w_j \log p(\tilde\theta_j|\tau_i)\right)  \\
		& - N \log \sum_{i=1}^N \exp \left(\log p(\tau_i) + \sum_{j=1}^M w_j \log p(\tilde\theta_j|\tau_i)\right)
	\end{aligned}
	\label{eqn:S-w-MC}
\end{equation}
When it is feasible to impose a uniform improper prior $p(\tau)
= C$ on the hyperparameter, such as in the case of a
sufficiently large number of groups in the hierarchy,
\eqref{eqn:S-w-MC} simplifies to \eqref{eqn:S-w-MC-uniform}:
\begin{equation}
	\begin{aligned}
		\hat S(\pmb{w}) &=  \sum_{i=1}^N \sum_{j=1}^M w_j \log p(\tilde\theta_j|\tau_i)  \\
		&- N \log \sum_{i=1}^N \exp \left(\sum_{j=1}^M w_j \log p(\tilde\theta_j|\tau_i)\right)
	\end{aligned}
	\label{eqn:S-w-MC-uniform}
\end{equation}
$\nabla_{\pmb{w}}\hat S(\pmb{w})$ is readily obtainable, either analytically or
algorithmically, and \eqref{eqn:w-arg-max-p-log-q} can be solved using gradient
ascent. Although $S(\pmb{w})$ is not concave in general, there is an obvious
initial guess $\pmb{w}=\pmb{1}$ in proximity of the global maximum; the
larger $M$, the closer is the guess to the maximum.

Let us illustrate the computation on a toy example. We use
model~\eqref{model:normal}
\begin{equation}
	\begin{aligned}
		p (\mu, \log \sigma) & \propto 1 \\
		y_j|\mu, \sigma & \sim \mathrm{Normal}(\mu, \sigma)
	\end{aligned}
	\label{model:normal}
\end{equation}
to infer the parameters of a single-dimensional normal distribution given
observations $\pmb{y}=\{ -1.33$, $-0.61$, $-0.20$, $0.34$, $0.71$, $1.23$,
$1.45$, $1.47$, $1.83$, $2.05\}$. The posterior is shown in blue in
Figure~\ref{fig:normal-posterior}.  Then, we draw 10 samples from the
predictive posterior, and condition~\eqref{model:normal} on the samples. The
new posterior is likely to be different, with the standard deviation of the \textit{mean} $\approx 0.33$. For $\tilde{\pmb{y}}=\{-2.77$,
$-1.80$, $-0.71$, $-0.62$, $0.31$, $0.38$, $0.43$, $0.70$, $1.66$, $2.6\}$, the
posterior is shown in green in Figure~\ref{fig:normal-posterior}. We then
compute, according to~\eqref{eqn:w-arg-max-p-log-q}
and~\eqref{eqn:S-w-MC-uniform}, the
weights (Figure~\ref{fig:normal-surrogate}), and condition
model~\eqref{model:normal} on the weighted sample, resulting in the posterior
(orange in Figure~\ref{fig:normal-posterior}) almost coinciding with the
original posterior.  In a hierarchical model, hyperparameters $\tau$ would
correspond to $\mu$ and $\sigma$, and group parameters $\pmb\theta$ to
$\pmb{y}$.
 
\begin{figure*}
\begin{subfigure}{0.38\linewidth}
	\centering
	\includegraphics[width=0.95\linewidth]{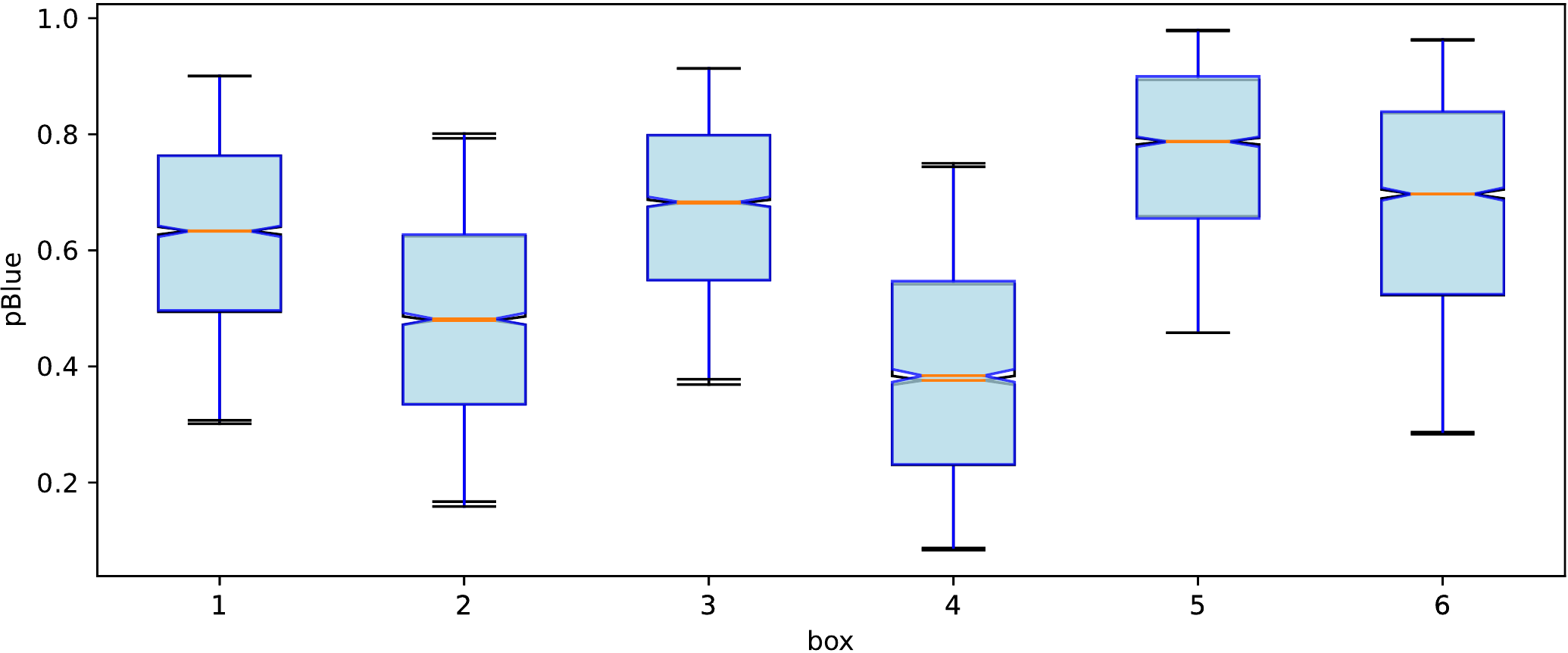}
	\caption{stump \& fungus vs. hierarchical}
	\label{fig:marbles-fit-sffit10}
\end{subfigure}
\begin{subfigure}{0.38\linewidth}
	\centering
	\includegraphics[width=0.95\linewidth]{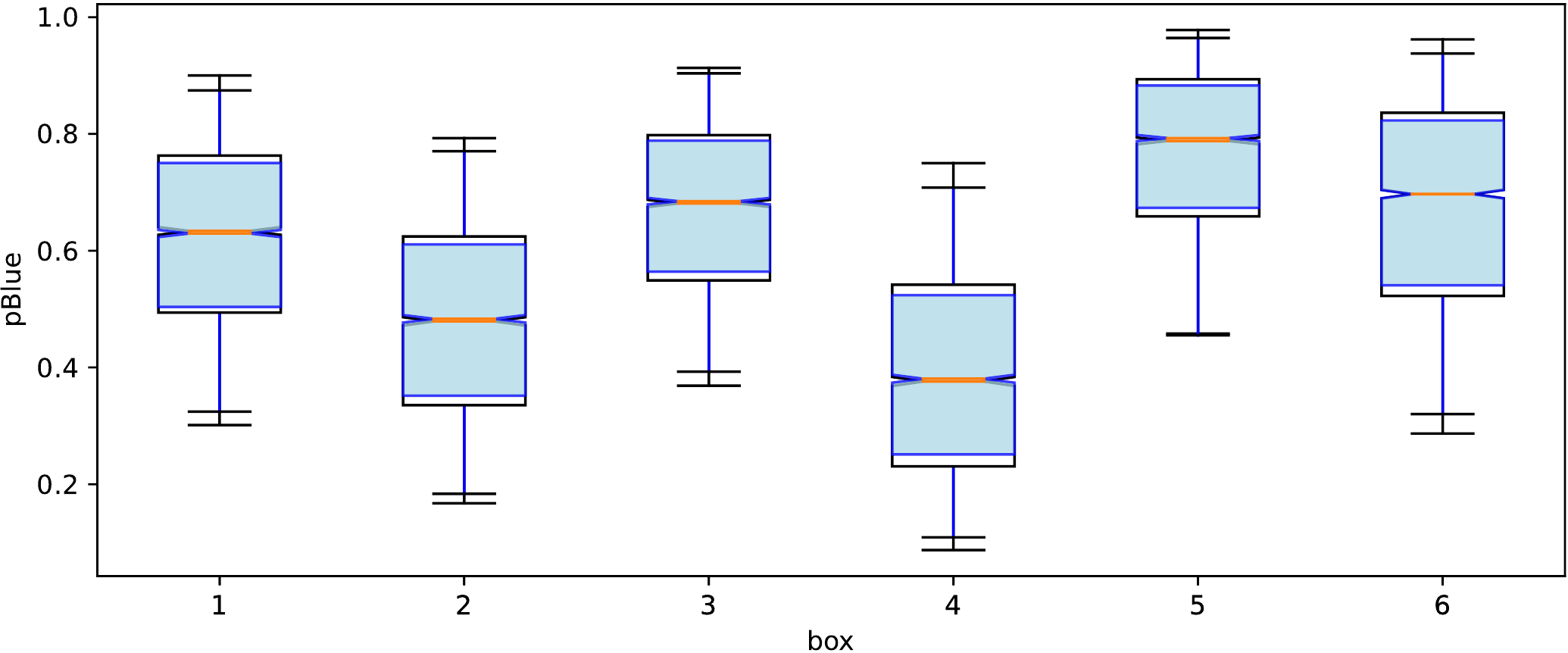}
	\caption{empirical Bayes vs. hierarchical}
	\label{fig:marbles-fit-ebfit}
\end{subfigure}
\begin{subfigure}{0.22\linewidth}
	\centering
	\includegraphics[width=0.95\linewidth]{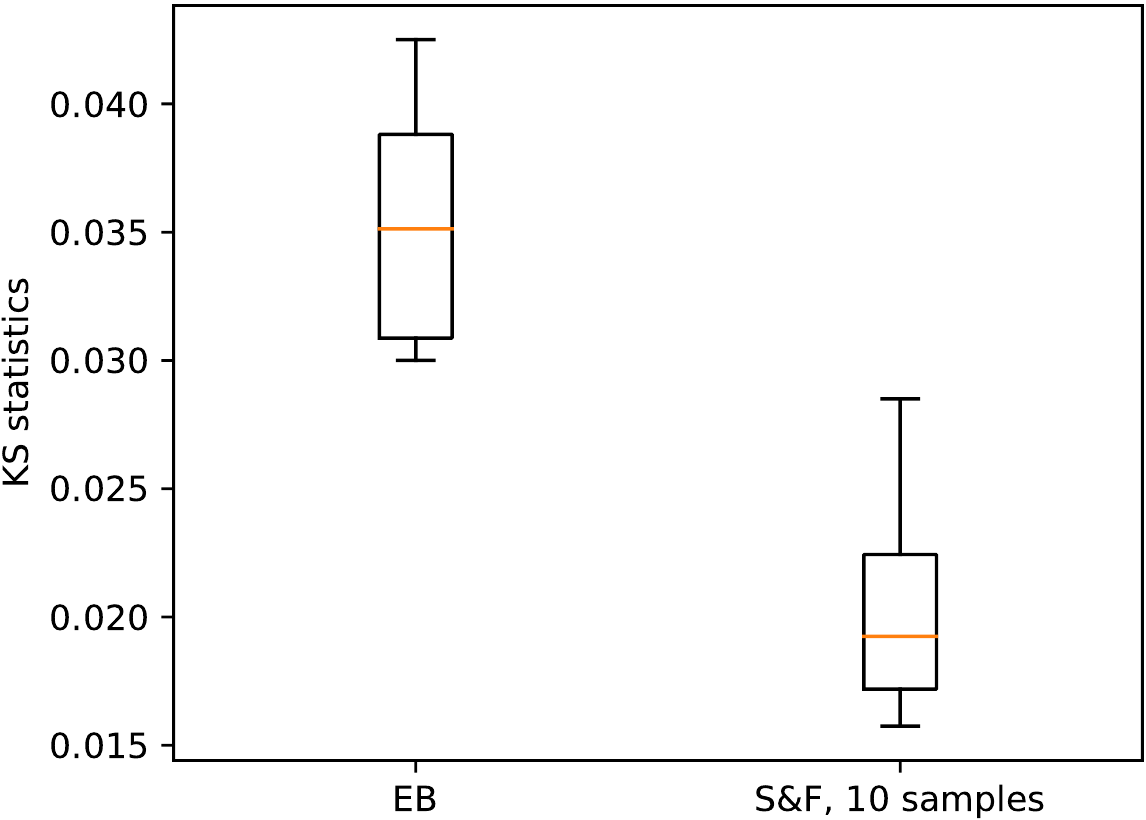}
	\caption{discrepancy}
	\label{fig:marbles-ks}
\end{subfigure}
	\caption{Marbles in boxes. Stump-and-fungus with 10 samples
	faithfully approximates the hierarchical posterior, while
	empirical Bayes results in a greater discrepancy.}
	\label{fig:marbles}
\end{figure*}

\section{Case Studies}
\label{sec:studies}

We evaluate the stump-and-fungus pattern on three case studies.
We begin with a basic synthetic case study ``Boxes With
Marbles'' (Section~\ref{sec:boxes}), which, while does not pose
computational challenges, allows us to show most aspects of
application of the pattern to a hierarchical model. We then
continue to a didactic study ``Tumor Incidence in Rats''
(Section~\ref{sec:rats}) which involves a large number of groups
in the hierarchy.  Finally, we apply the pattern to
``Educational Attainment in Secondary Schools''
(Section~\ref{sec:attain}), an elaborated cross-classified model
based on data used in a real-world sociological research. We
implemented in the studies in Infergo~\citep{T19} and fit models
using HMC~\citep{N12} or NUTS~\citep{HG11}. We used gradient
ascent with momentum for computation of sample weights.  The
data and source code for the case studies are provided in the
supplementary material.

\subsection{Boxes With Marbles}
\label{sec:boxes}

This case study is inspired by an introductory example
in~\citet{M20}. Boxes are filled with marbles from a bag with a
mix of blue and white marbles. There are 6 boxes, with 4 marbles
in each box.  Marbles are drawn from the boxes, with
replacement, and observed. The number of blue marbles in each
box are to be inferred. The problem is formalized by
model~\eqref{model:marbles}:
\begin{equation}
	\begin{aligned}
		p_0 &\sim \mathrm{Uniform}(0, 1) \\
		p_i|p_0 &\sim \mathrm{Beta}\left(4p_0, 4\left(1-p_0\right)\right) \\
		y_j|p_{b_j} & \sim \mathrm{Bernoulli}\left(p_{b_j}\right)
	\end{aligned}
	\label{model:marbles}
\end{equation}
where $p_0$ is the proportion of blue marbles in the bag, $p_i$ --- in the
$i$th box, $y_j$ is 1 if the $j$th drawn marble was blue, 0 otherwise, and
$b_j$ is the box from which the $j$th marble was drawn. 

We fit 
\begin{itemize}
\item the hierarchical model;
\item the empirical Bayes model in which $p_0$ is fixed to the posterior mean of $p_0$ in the hierarchical model;
\item 6 stump-and-fungus models, with each box as the fungus, in
turn, using 10 samples from the parameter posterior for the
stump.
\end{itemize}
Figure~\ref{fig:marbles} compares the posteriors.
Figure~\ref{fig:marbles-fit-sffit10} shows the posterior
distributions of $p_i$ from the hierarchical model (white) and each of
the stump-fungus models (light blue), overlaid. Similarly,
Figure~\ref{fig:marbles-fit-ebfit} compares the posteriors of
the hierarchical and the empirical Bayes models. One can observe
that stump-and-fungus results in a better approximation of the
posterior than empirical Bayes. Figure~\ref{fig:marbles-ks}
visualizes discrepancy between the posteriors with
the distribution of Kolmogorov-Smirnov statistics of $p_i$ over
boxes, confirming the advantage of stump-and-fungus.

\subsection{Tumor Incidence in Rats}
\label{sec:rats}

\begin{figure*}
	\begin{subfigure}{0.325\linewidth}
		\centering
		\includegraphics[width=0.95\linewidth]{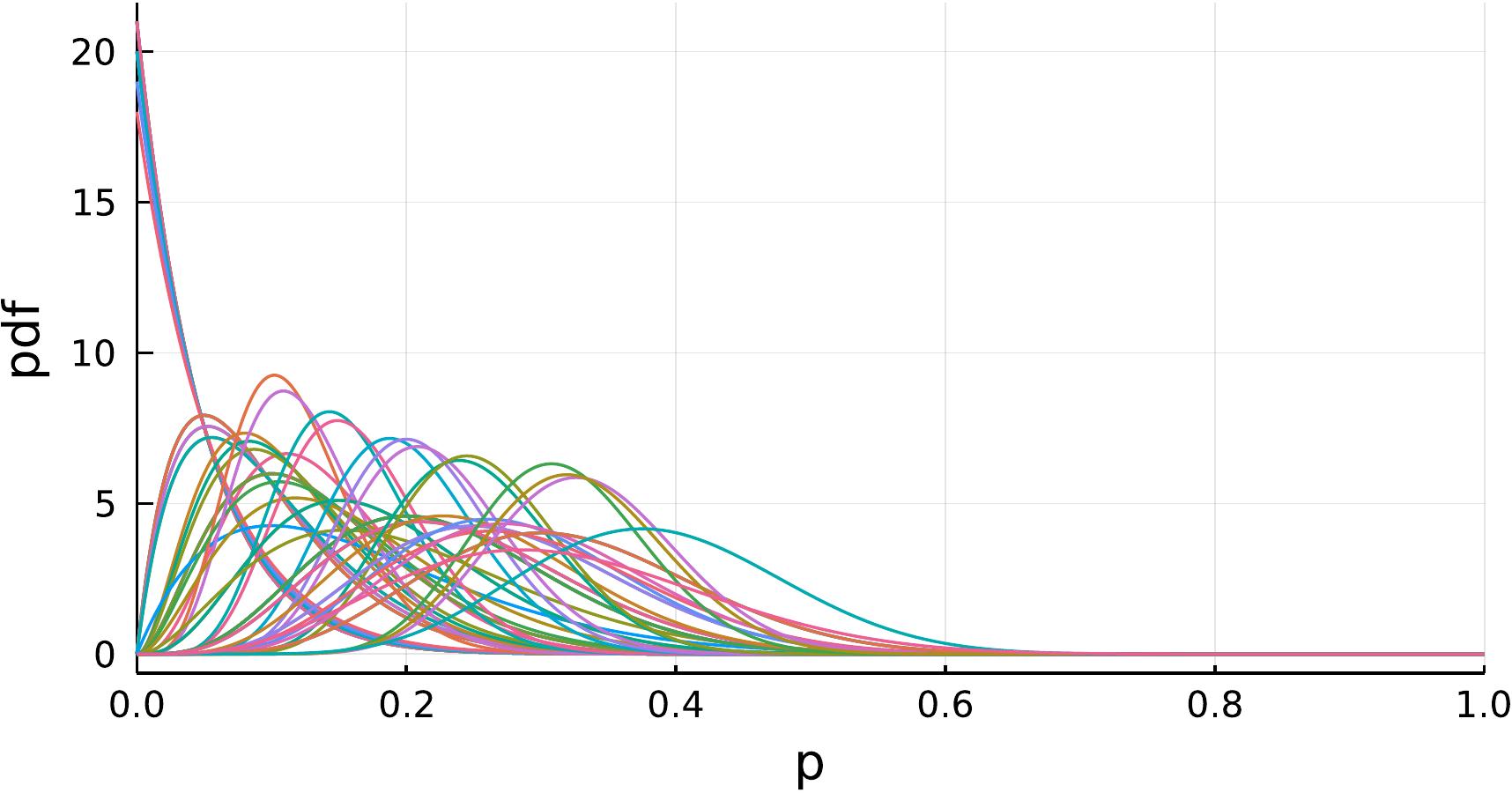}
		\caption{Unpooled models}
		\label{fig:separate}
	\end{subfigure}
	\begin{subfigure}{0.325\linewidth}
		\includegraphics[width=0.95\linewidth]{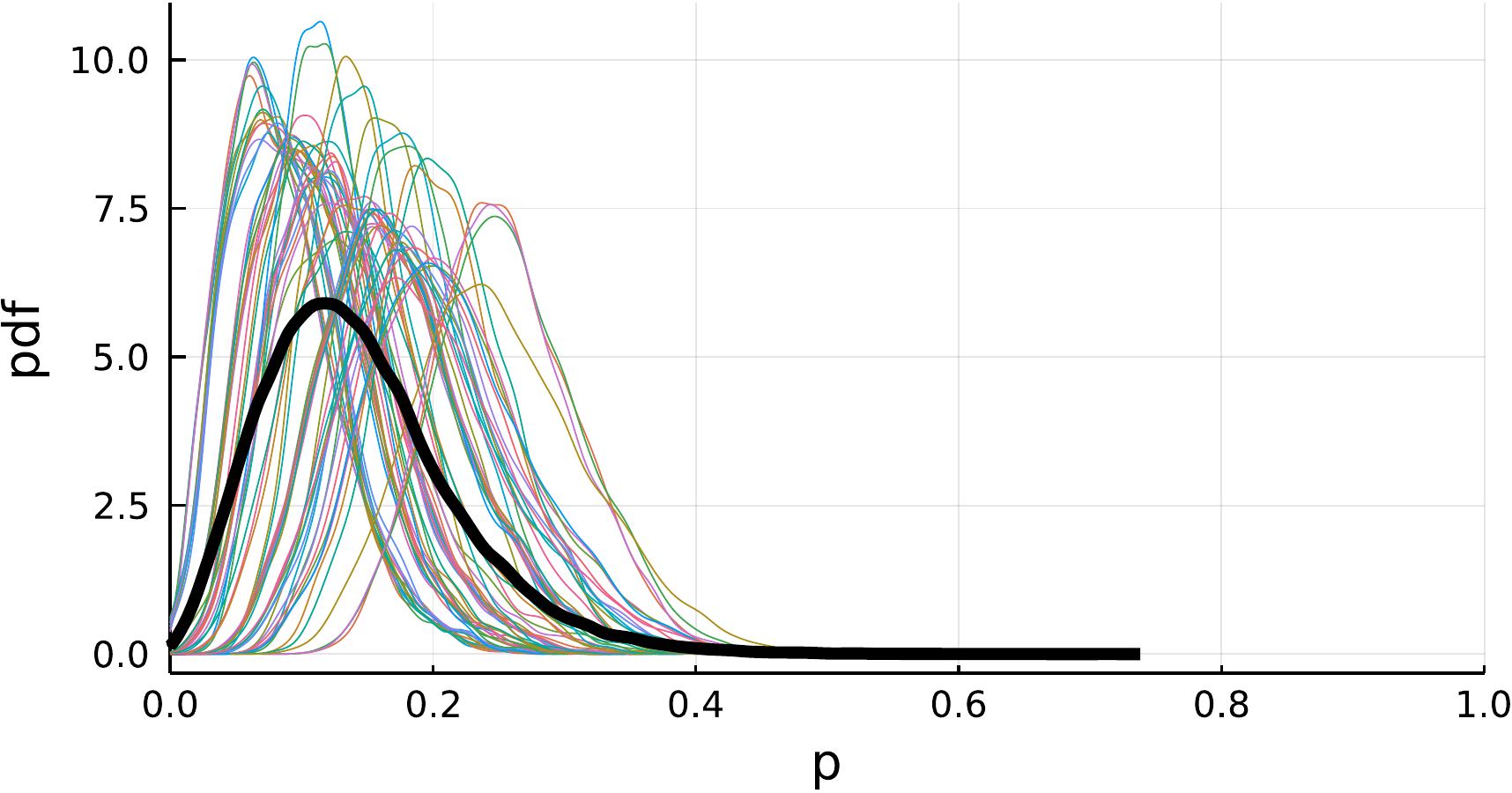}
		\caption{Hierarchical model}
		\label{fig:tree}
	\end{subfigure}
	\begin{subfigure}{0.325\linewidth}
		\includegraphics[width=0.95\linewidth]{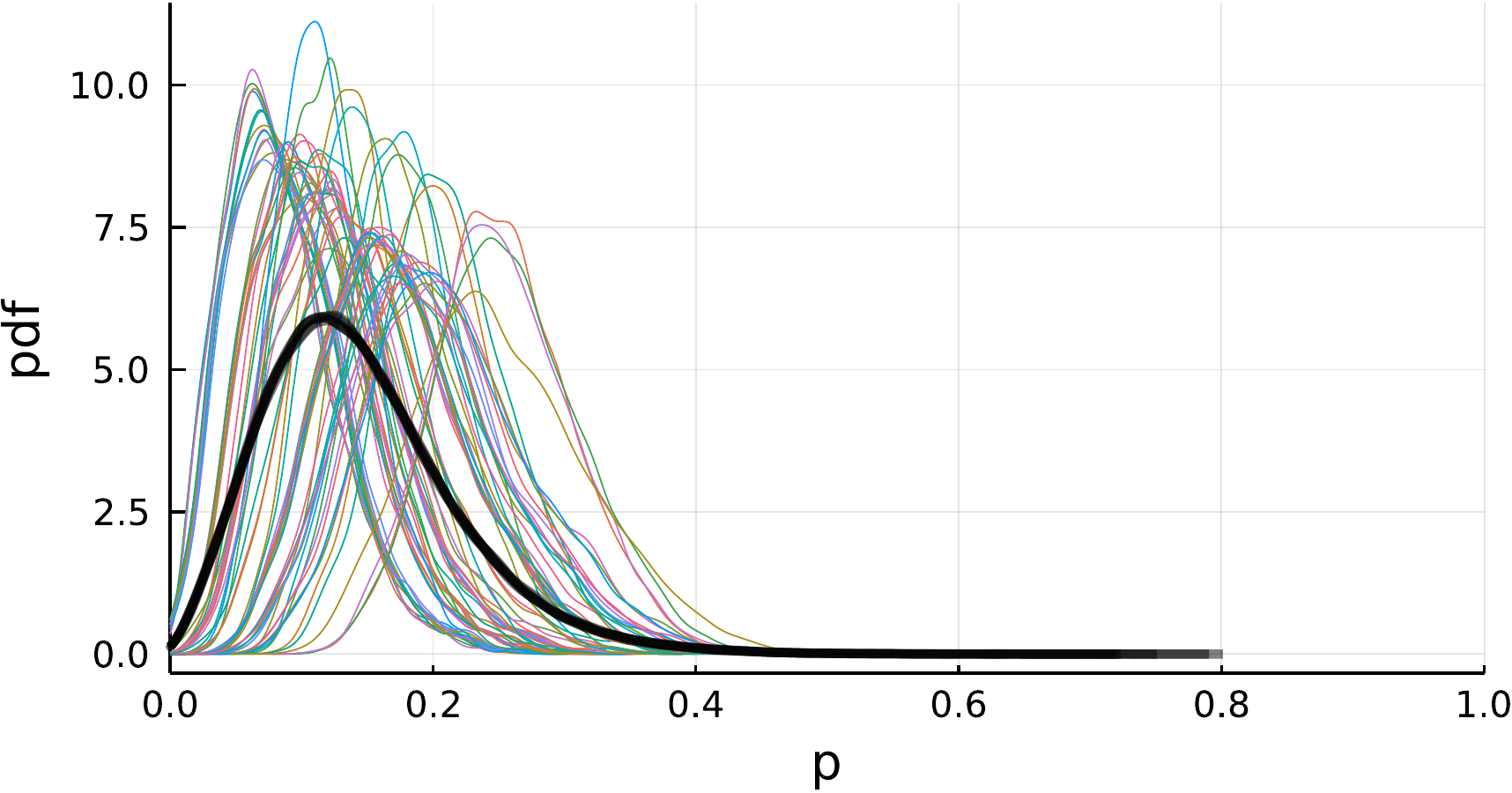}
		\caption{Stump-and-fungus models}
		\label{fig:fungi}
	\end{subfigure}
	\caption{Tumor incidence in rats. The posteriors are the same in
	hierarchical and stump-and-fungus model, except
	for small discrepancies apparently caused by finite sample
	size approximation. Colorful lines are posterior
	distributions of $p$ for each of 71 experiments. 
	Black lines are posterior distributions of $p$ for
	a future experiment.}
	\label{fig:tree-vs-fungi}
\label{fig:rats}
\end{figure*}

In this case study, based on \citet{T82} and discussed in
\citet[Chapter 5]{GCS+13}, data on tumor incidence in rats in $N=70$
laboratory experiments is used to infer tumor incidence based on
outcomes of yet another experiment. A different number of rats
$n_i$ was involved in each experiment, and the number of tumor
cases $y_i$ was reported. An application of the stump-and-fungus
pattern in this case would be to replace the original data set of
size $N\times 2$ with a `stump' of a smaller size $M\times 2$ for
assessment of outcomes of future experiments. The problem is formalized by
model~\ref{model:rats-hier}:
\begin{equation}
	\begin{aligned}
		\alpha, \beta & \sim \mathrm{Prior} \\
		p_i|\alpha, \beta & \sim \mathrm{Beta}(\alpha, \beta) \\
		y_i|p_i & \sim Binomial(n_i, p_i) 
	\end{aligned}
	\label{model:rats-hier}
\end{equation}
where the non-informative prior distribution is set such that
$p(\alpha, \beta) \propto (\alpha +
\beta)^{-5/2}$~\citep[Section~5.3]{GCS+13}. In addition, each
group can be fit an `unpooled' Beta-Binomial
model~\eqref{model:rats-unpooled}: 
\begin{equation}
	\begin{aligned}
		p_i & \sim \mathrm{Beta}(1, 1) \\
		y_i|p_i & \sim \mathrm{Binomial}(n_i, p_i) 
	\end{aligned}
	\label{model:rats-unpooled}
\end{equation}
We fit 
\begin{itemize}
\item a separate `unpooled' model for each experiment;
\item the hierarchical model;
\item 71 stump-and-fungus models, with each experiment as the
fungus, in turn, using 10 samples from the parameter posterior
for the stump.
\end{itemize}
Since \citet[Section 5.3]{GCS+13} give a detailed account of
empirical Bayes vs. hierarchical model, we omit this comparison
here.

Inference on model~\eqref{model:rats-hier} can be performed
efficiently thanks to summarization of $n$ observations from
$\mathrm{Bernoulli}(p)$ as a single observation from
$\mathrm{Binomial}(n, p)$. In general however, the use of a
hierarchical model would require carrying all observations of
all previous experiments for learned inference on findings of a
new experiment. 

Figure~\ref{fig:tree-vs-fungi} shows the posterior distributions
for $p$ inferred on 71 unpooled models (Figure~\ref{fig:separate}), on
the hierarchical model (Figure~\ref{fig:tree}) and through
$71$ applications of stump-and-fungus
(Figure~\ref{fig:fungi}). One can observe that the
posteriors obtained on both hierarchical and
stump-and-fungus models appear to be the same, except for
small discrepancies apparently caused by finite sample size
approximation. Compared to the unpooled posteriors,
significant shrinkage takes place. Table~\ref{tab:rats-times}
shows the inference times for each of the models.
Inference on a single group with the stump-and-fungus model
takes less than 10\% of time on the hierarchical
model.

\begin{table}[h]
	\centering
	\caption{Tumor incidence in rats: inference times
	for 1000 burn-in and 5000 posterior samples, 10 stump samples. }
	\label{tab:rats-times}
	\setlength\tabcolsep{2pt}
	\footnotesize
	\begin{tabular}{r|c|c|c|c} 
		\textit{Model} & Hierarchical & Separate & Stump \& Fungus & Weights \\ \hline
		\textit{Time, sec}  &  93     &   2.1    & 7.2             & 15$\pm$1.1
	\end{tabular}
\end{table}

\subsection{Educational Attainment in Secondary Schools}
\label{sec:attain}

This study is based on \citet{P91} and explores educational attainment of
children in Scotland. There are two real-valued predictors: social class scale
\texttt{CC} and verbal reasoning score \texttt{VRQ} on entry to the secondary
school, three hierarchies: by secondary school(\texttt{SID}), by gender
(\texttt{SEX}), and by primary school (\texttt{PID}), and a real valued target
--- attainment score \texttt{ATTAIN} at the age of 16. There are 3,435
children, 148 primary schools, and 19 secondary schools in the data set. An
application of stump-and-fungus problem would be to provide a new secondary
school with a small `stump', instead of the full data set, for attainment
assessment. The problem can be formalized as model~\eqref{model:attain}, 
a cross-classified hierarchical linear regression model:
\begin{equation}
	\begin{aligned}
		& \mu_{\beta}, \log \sigma_{\beta}, \mu_{\sigma}, \log \sigma_{\sigma}  \sim \mathrm{Uniform}(-\infty, \infty) \\
		& \beta|\mu_{\beta}, \sigma_{\beta}  \sim \mathrm{Normal}(\mu_{\beta}, \sigma_{\beta}) \\
		& \log \sigma|\mu_{\sigma}, \sigma_{\sigma}  \sim \mathrm{Normal}(\mu_{\sigma}, \sigma_{\sigma}) \\
		& p(y_i|\beta, \sigma) = \prod_h \mathrm{Normal}(y_i\,|\,\beta_{h g_{ih}}\!\cdot x_i, \sigma_{h g_{ih}})
	\end{aligned}
	\label{model:attain}
\end{equation}
Here, $h$ is the hierarchy index, $x_i$ and $y_i$ are the
predictors and the target of the $i$th pupil, $g_{ih}$ is the
group index of the $i$th pupil in the $h$th hierarchy.
$\mu_{\beta}, \log \sigma_{\beta}, \mu_{\sigma}, \log
\sigma_{\sigma}$ are the hyperparameters of the model, and
$\beta, \sigma$ are the group parameters; in total, the model
has 700 real-valued parameters, of which 18 are hyperparameters. 

\begin{figure*}
\begin{subfigure}{\linewidth}
	\centering
	\includegraphics[width=0.9\linewidth]{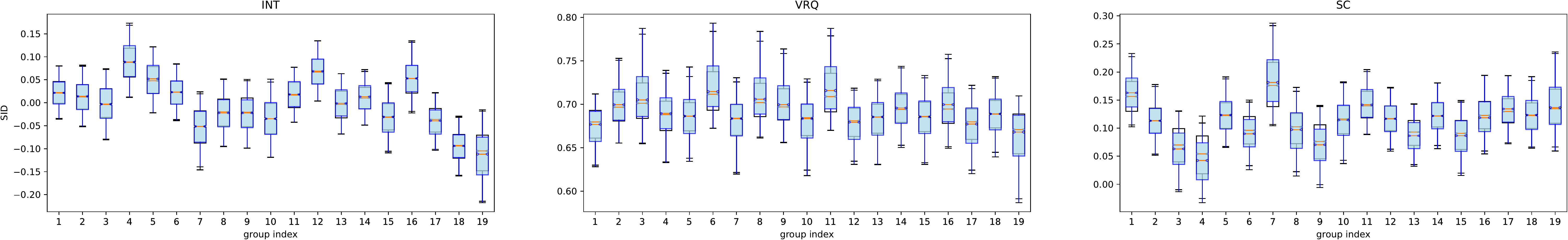}
	\caption{stump \& fungus, 10 samples vs. hierarchical}
	\label{fig:attain-fit-sffit10}
\end{subfigure}

\begin{subfigure}{\linewidth}
	\centering
	\includegraphics[width=0.9\linewidth]{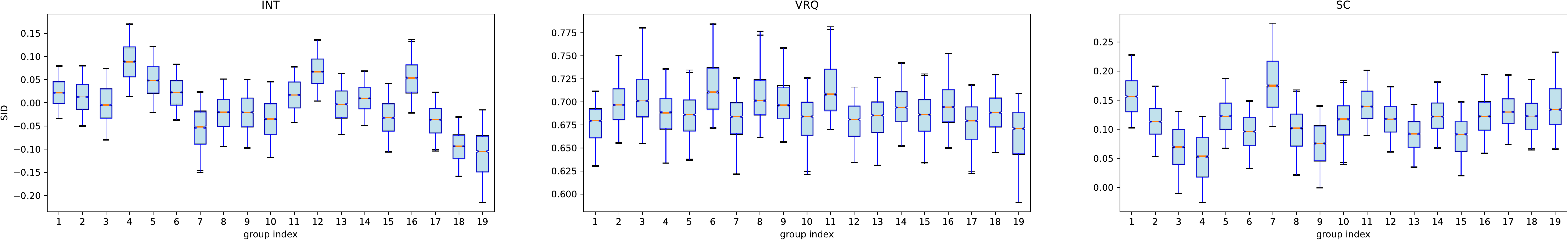}
	\caption{stump \& fungus, 20 samples vs. hierarchical}
	\label{fig:attain-fit-sffit20}
\end{subfigure}

\begin{subfigure}{\linewidth}
	\centering
	\includegraphics[width=0.9\linewidth]{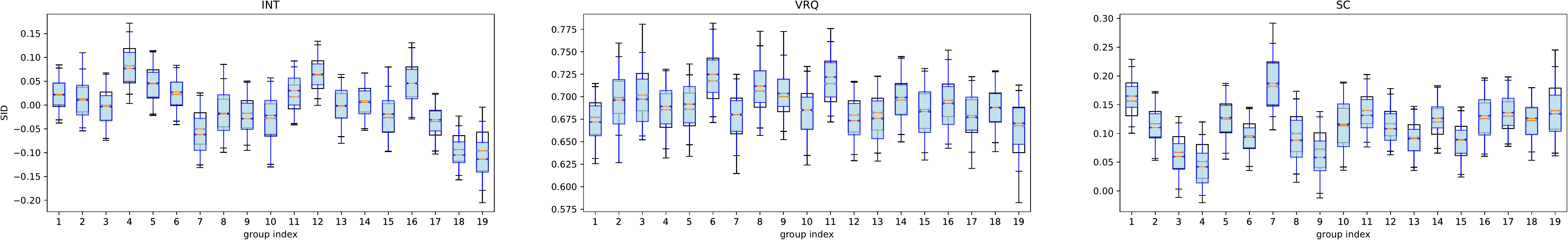}
	\caption{empirical Bayes vs. hierarchical}
	\label{fig:attain-fit-ebfit}
\end{subfigure}

\begin{subfigure}{\linewidth}
	\centering
	\includegraphics[width=0.5\linewidth]{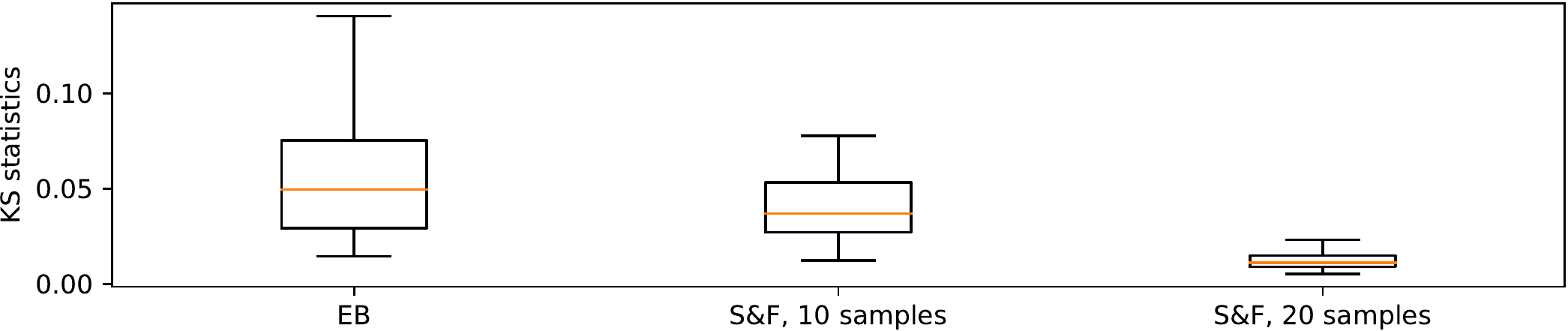}
	\caption{discrepancy}
	\label{fig:attain-ks}
\end{subfigure}

	\caption{Educational attainment. Stump-and-fungus with both 10 and 20 samples
	 approximate the hierarchical posterior better than empirical Bayes.}
	\label{fig:attain}
\end{figure*}

We fit 
\begin{itemize}
\item the hierarchical model;
\item the empirical Bayes model in which the hyperparameters are
	fixed to their posterior means in the hierarchical model;
\item 19 stump-and-fungus models, with each secondary school as the fungus, in
	turn, using 10 and 20 samples from the parameter posterior for the stump.
	Since parameters in each hierarchy are mutually independent given
	hyperparameters, a separate weight is assigned to each component of a
	sample, rather than to the sample as a whole.
\end{itemize}
Figure~\ref{fig:attain} compares the posteriors.
Figures~\ref{fig:attain-fit-sffit10}
and~\ref{fig:attain-fit-sffit20} show the posterior distributions of
$p_i$ from the hierarchical model (white) and each of the
stump-fungus models (light blue) for 10 and 20 samples,
correspondingly. Similarly, Figure~\ref{fig:attain-fit-ebfit}
compares the posteriors of the hierarchical and the empirical
Bayes models. One can observe that stump-and-fungus results in a
better approximation of the posterior than empirical Bayes.
Figure~\ref{fig:attain-ks} visualizes discrepancy between the
posteriors with the distribution of Kolmogorov-Smirnov
statistics of the group parameters over the secondary schools,
confirming the advantage of stump-and-fungus.  Table~\ref{tab:attain-times}
shows the inference times for each of the models.
Inference on a single group with the stump-and-fungus model
takes less than 25\% of time on the hierarchical
model.

\begin{table}[h]
	\centering
	\caption{Educational attainment: inference times
	for 1000 burn-in and 5000 posterior samples, 10 and 20 stump samples. }
	\label{tab:attain-times}
	\setlength\tabcolsep{2pt}
	\footnotesize
	\begin{tabular}{r|c|c|c|c|c|c}
		\textit{Model} & Hierar- & Empirical & \multicolumn{2}{c|}{Stump \& Fungus }&\multicolumn{2}{c}{Weights} \\ 
					   & chical  & Bayes     & 10 & 20 & 10 & 20 \\ \hline
		\textit{Time, sec} & 908 & 860 & 180$\pm$12 & 224$\pm$15 & 236$\pm$30 & 490$\pm$55
    \end{tabular}
\end{table}

\section{Related Work}
\label{sec:related}

The importance of learning from data is well
appreciated in probabilistic programming. Along with empirical
Bayes, applicable to probabilistic programming as well as to
Bayesian generative models in general, probabilistic-programming
specific approaches were proposed. One possibility is to induce
a probabilistic program suited for a particular data
set~\citep{LJK10, PW14, P18, HSG11}.  A related but different
research direction is inference compilation~\citep{LBW17,BSB+19},
where the cost of inference is amortized through learning
proposal distributions from data. Another line of research is
concerned by speeding up inference algorithms by tuning them
based on training data~\citep{ETK14,MSH+18}.  Our approach to
learning from data in probabilistic programs is different in
that it does not require any particular implementation of
probabilistic programming to be used, nor introspection into the
structure of probabilistic programs or inference algorithms.
Instead, the approach uses inference in ubiquitously adopted
hierarchical models for training, and conditioning on
observations for incorporation of training outcomes in
inference. 

Approximation of a large sample set by a small weighted subset
bears similarity to Bayesian
coresets~\citep{HCB16,CB19,MXM+20,ZKK+21} --- a family of
approaches aiming at speeding up inference with large datasets.
A Bayesian coreset is a small weighted subset of the original
large dataset, with the promise that inference on the coreset
yields the same or approximately the same posterior. However,
there are significant differences between Bayesian coresets and
the setting in this work. First, in Bayesian coresets, the
posterior is unknown when the coreset is constructed. In the
stump-and-fungus pattern, the posterior distribution of the
hyperparameter is known (as  a Monte Carlo approximation) before
the samples are selected and the weights are computed.  In
particular, \cite{CB19} minimize the KL divergence between the
approximate and the exact posterior, while
\eqref{eqn:w-arg-max-p-log-q} in this work minimizes the
complementary KL divergence between the exact posterior and its
stump-and-fungus approximation.  This facilitates a simpler
formulation of divergence minimization. Second, careful
selection of samples to be included in the coreset is
necessitated by high dimensionality of data in the
dataset~\citep{MXM+20}. However, even in elaborated hierarchical
Bayesian models the group parameters are low-dimensional, and
interdependencies between multiple hierarchies in
cross-classified models are assumed to be negligible. Because of
that, a random draw from the parameter posterior suffices for
stump-and-fungus even if it might not work in a
higher-dimensional setting. 

\section{Discussion}
\label{sec:discussion}

We presented a probabilistic programming pattern for Bayesian
learning from data.  The stump-and-fungus pattern is easy to
understand and simple to implement. When applied to hierarchical
generative models, the pattern allows to realize learning from
data within the Bayesian paradigm, and results in significant
improvement in inference performance on new groups. Even if
certain aspects of the pattern can be improved, the pattern is
applicable in its current form to real-world statistical
studies. 


\bibliographystyle{named}
\bibliography{refs}

\end{document}